\documentclass[10pt,twocolumn,letterpaper]{article}

\usepackage{cvpr}              

\usepackage{graphicx}
\usepackage{amsmath}
\usepackage{amssymb}
\usepackage{booktabs}
\usepackage{bm}
\usepackage{booktabs}
\usepackage{colortbl}

\usepackage[pagebackref,breaklinks,colorlinks]{hyperref}

\usepackage[capitalize]{cleveref}
\crefname{section}{Sec.}{Secs.}
\Crefname{section}{Section}{Sections}
\Crefname{table}{Table}{Tables}
\crefname{table}{Tab.}{Tabs.}

\def\cvprPaperID{*****} 
\def\confName{CVPR}
\def\confYear{2023}
\definecolor{Gray}{gray}{0.9}

\begin{document}

\title{A Dual-level Detection Method for Video Copy Detection}

\author{Tianyi Wang\textsuperscript{\rm 1}\thanks{Equal contribution.}, Feipeng Ma\textsuperscript{\rm 1,2}\footnotemark[1],  Zhenhua Liu\textsuperscript{\rm 1}\footnotemark[1],  Fengyun Rao\textsuperscript{\rm 1}\thanks{Corresponding author.}\\
\textsuperscript{\rm 1}WeChat of Tencent, 
\textsuperscript{\rm 2}University of Science and Technology of China  \\
{\tt\small tyewang@tencent.com, mafp@mail.ustc.edu.cn, edinliu@tencent.com, fengyunrao@tencent.com}}


\maketitle


\begin{abstract}
With the development of multimedia technology, Video Copy Detection has been a crucial problem for social media platforms. Meta AI hold Video Similarity Challenge on CVPR 2023 to push the technology forward. In this paper, we share our winner solutions on both tracks to help progress in this area. For Descriptor Track, we propose a dual-level detection method with Video Editing Detection (VED) and Frame Scenes Detection (FSD) to tackle the core challenges on Video Copy Detection.  Experimental results demonstrate the effectiveness and efficiency of our proposed method.
Code is available at \url{https://github.com/FeipengMa6/VSC22-Submission}.
   
\end{abstract}

\section{Introduction}
\label{sec:intro}
In the past decade, the development of information technology has led to a shift in the main carrier of information from text to images and then to videos. Moreover, with the rise of User-generated Content (UGC), the producer of information has shifted from Occupationally-generated Content (OGC) to UGC. As a result, a large number of videos have emerged on social media platforms and have been widely shared, leading to the increasingly important and challenging problems of video copyright protection. 
The core challenges of video copy detection are twofold: effective video descriptors and computational costs. Copied videos often involve edited portions, making it difficult for general visual models to differentiate between copied and original content. A powerful model must be capable of discriminating between videos, even when significant editing has taken place. Additionally, the cost of time and resources required to process each query video and identify the most similar reference video is a significant concern, necessitating the development of efficient and cost-effective methods.
\begin{figure}[tbp]
    \centering
    \includegraphics[width=0.48\textwidth]{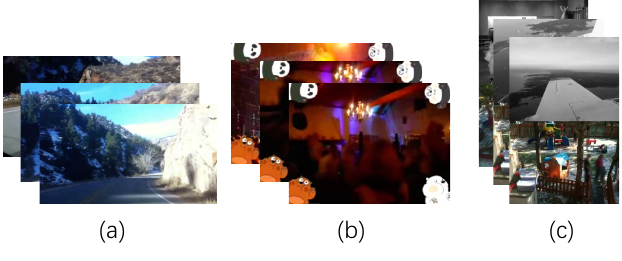}
    \caption{Three typical situations of query videos. (a) is an unedited video, which is the most of query videos. (b) is a copied video with general editing operation. (c) is a copied video with multiple scenes in each frame. }
    \label{fig:abs}
\end{figure}

In this paper, we summarize our proposed method for Meta AI Video Similarity Challenge, which tackle the core challenges of video copy detection through a dual-level detection method. In Fig.~\ref{fig:abs}, there are three typical situations of query videos, including unedited video, copied video with general editing operation and copied video with multiple scenes in each frame. Our proposed dual-level detection method first identify if the video has been edited in video-level. For unedited videos, we use random vectors with small norm as their descriptors. What's more, we replace the bias term of these descriptors  with a negative value during score normalization. For edited videos, we notice that it is necessary to deal with the situation that multiple scenes are concatenated along edge. We adopt traditional image processing method to detect and split the scenes in one frame. With our dual-level detection method, we can reduce the storage cost for unedited videos and improve the efficiency and accuracy of retrieval.

\begin{figure*}[htbp]
    \centering
    \includegraphics[width=0.9\textwidth]{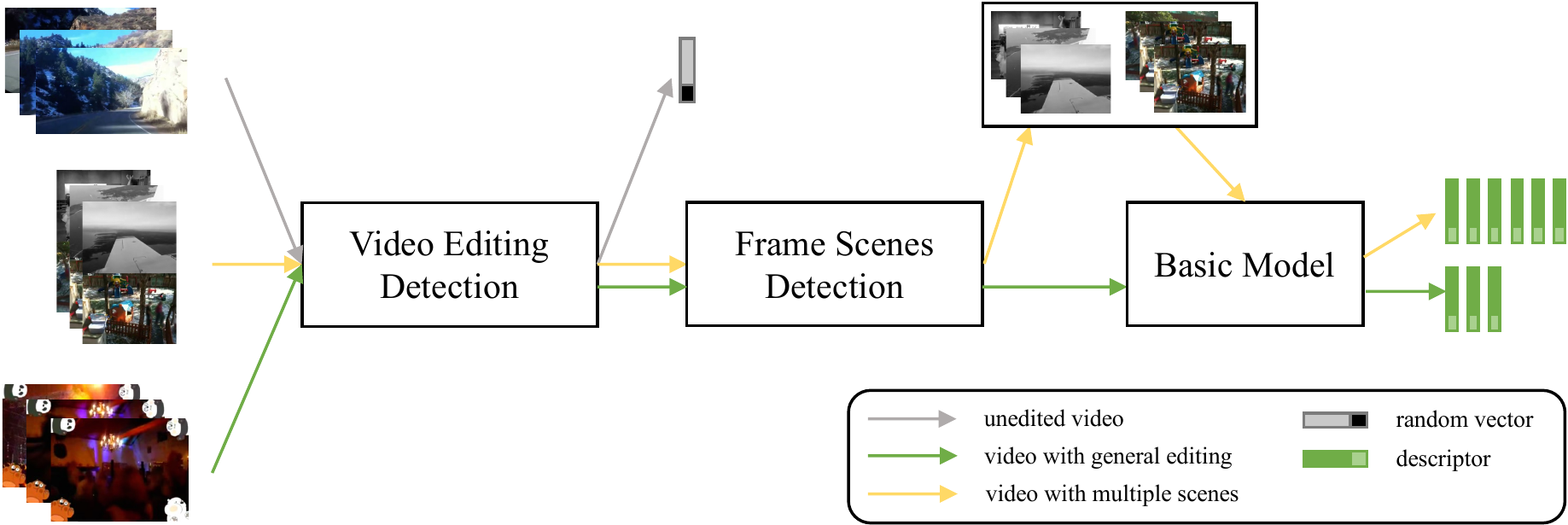}
    \caption{Overview of our proposed pipeline. For unedited video, we directly use a random vector with samll norm and negative bias term. For video with general editing, we extract features of each frame by our pre-trained basic model as descriptors. For video with multiple scenes, we detect and split the scenes in each frame, then use basic model to generate descriptors. }
    \label{fig:framework}
\end{figure*}

The main contributions are summarized as follows:
\begin{itemize}
    \item We propose a dual-level detection method for Descriptor Track, which detects edited videos at video-level and multiple scenes at frame-level. With the dual-level detection, we can reduce the computational cost and improve the performance. 
    \item The proposed method achieve outstanding performance on Meta AI Video Similarity Challenge and we got second prize on Descriptor Track. Our ablation study shows the effective of each module.
\end{itemize}





\section{Method}
In this section, we first introduce the design of our basic model. Then we explain the details of our proposed dual-detection method including video editing detection, frame scenes detection and score filter normalization.

\subsection{Basic Model}
We train a basic model to extract descriptors for video copy detection.
There are two potential types of descriptors that can be employed in video copy detection: video-level features or frame-level features. Given that our objective is not only to discriminate copied videos but also to identify copied portions between query and reference videos, we selected frame-level features as the video descriptor. Therefore, we adopt image transformer~\cite{dosovitskiy2020image,liu2021swinv2} as backbone. We follow SSCD~\cite{pizzi2022self} to train our basic model in a self-supervised manner. As SSCD uses, we combine SimCLR~\cite{chen2020simple} method with entropy loss~\cite{sablayrolles2018spreading}.

\noindent \textbf{The InfoNCE Loss.} We use the InfoNCE loss in SimCLR, which is softmax cross-entropy loss with temprature. The loss function is formulated as follow:
\begin{equation}
    L_{\rm InfoNCE} = -\frac{1}{|P|} \sum_{i,j \in P} \log \frac{\exp (\cos(z_i,z_j)/\tau)}{\Sigma_{k \neq i} \exp (cos(z_i,z_k)/\tau) }
\end{equation}
Where $P$ is the set of positive pairs, $z_i$ represents the descriptor,  $\tau$ is the temperature.

\noindent \textbf{The Entropy Loss.} We follow SSCD using the entropy loss proposed in~\cite{sablayrolles2018spreading}. The loss function is formulated as:
\begin{equation}
    L_{\rm KoLeo} = -\frac{1}{N} \sum_{i=1}^{N} \log (\min_{j\neq i} ||z_i - z_j||)
\end{equation}
Where $N$ is the size of training set.

\noindent \textbf{The Final Loss.} The final loss function is:
\begin{equation}
    L = L_{\rm InfoNCE} + \lambda L_{\rm KoLeo}
\end{equation}
Where $\lambda$ is the weights of Entropy Loss term.
Cause our training process of basic model is based on SSCD~\cite{pizzi2022self}, more details can be found in this paper.
\subsection{Video Editing Detection}
\label{sec:video_editing_detection}
To address the issue of high computational costs in video copy detection and provide an efficient solution, we propose a straightforward method to identify edited videos before generating frame-level descriptors. We observe that videos with copies are often edited, incorporating techniques such as blending, blurring, rotations, and other manipulations. This is due to the fact that copied videos must frequently merge multiple clips, necessitating additional editing operations. By filtering out videos that have not been edited among the query videos, we can reduce computational costs.
To achieve this goal, we have developed a model capable of discriminating between edited and unedited videos. Since editing operations can be viewed as strong augmentations, we aim to identify videos with such augmentations using a binary classification approach. We utilize CLIP~\cite{radford2021learning} to extract frame features without any post-processing and feed these features into RoBERTa~\cite{liu2019roberta}. And we employ the class token to calculate cross entropy.
We find that the edited video detection can achieve high accuracy, and using a small value $\alpha$ as threshold can filter most of unedited videos. For the unedited video, we use a random vector with very small value as descriptor. This processing can reduce the storage cost for query videos and speed up searching.

\subsection{Frame Scenes Detection}
\begin{figure}[htbp]
    \centering
    \includegraphics[width=0.48\textwidth]{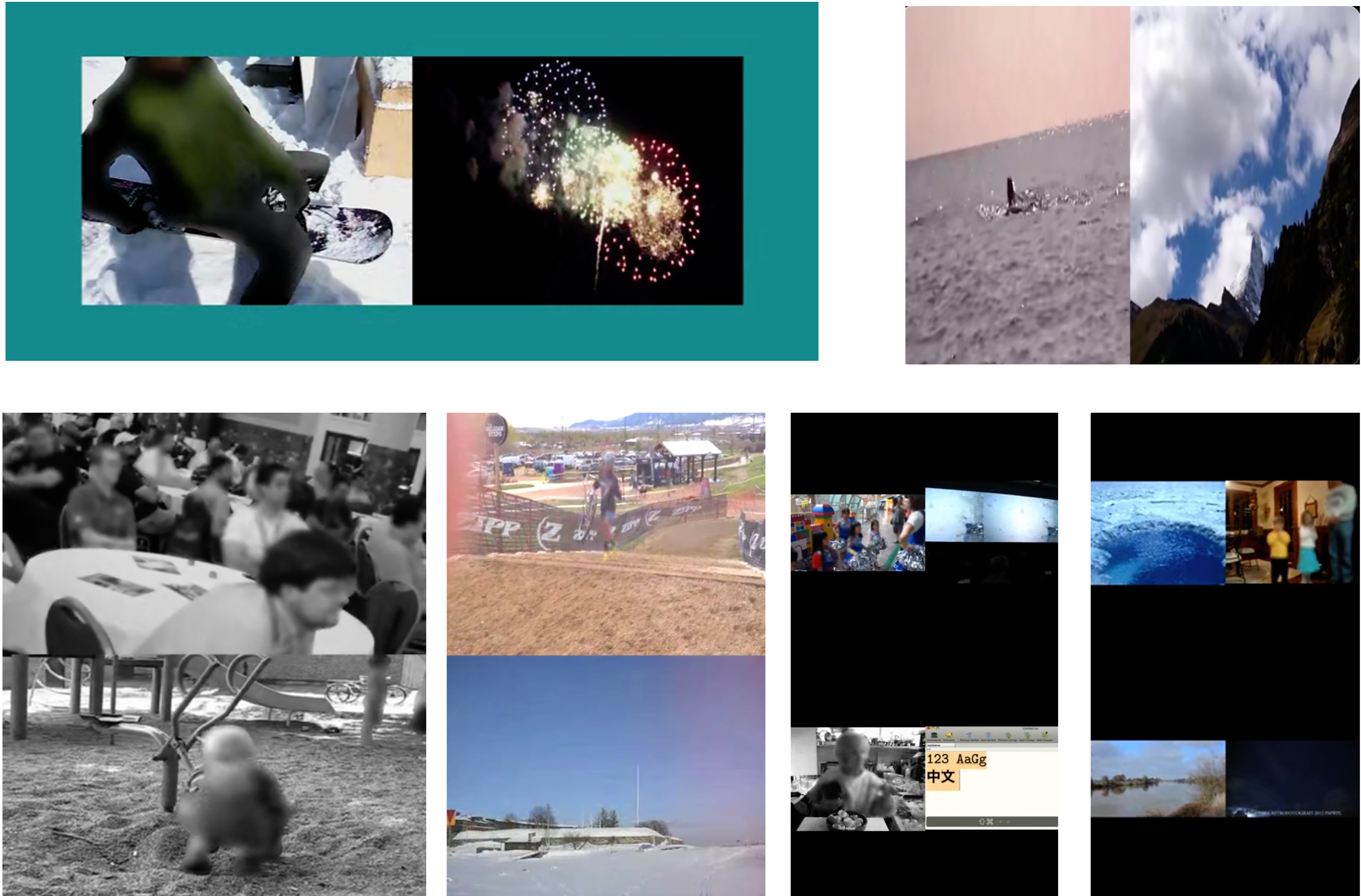}
    \caption{Multiple scenes in one frame.}
    \label{fig:multiple_scenes}
\end{figure}

We notice that stacking multiple scenes in one frames is an obvious augmentation of copied videos, and simple traditional image processing method can deal with this situation well.
Due to the continuity of the video, the combination of multiple scenes in one frames are limited. As shown in Fig.~\ref{fig:multiple_scenes}, scenes are usually concatenated along one side and one frame usually has an even number of scenes, most of them have two or four scenes. 

\subsection{Score Filter Normalization}
We follow~\cite{pizzi2022self,douze20212021} using similarity normalization in our evaluation. It introduce a background image dataset and only queries whose similarity score with reference is much higher than images in background dataset will have high scores. Based on it, we modify the integrated bias to suppress the score of unedited videos. In Sec.~\ref{sec:video_editing_detection}, we use a random vector with very small value as descriptor for unedited video, but scores of these videos are clustered around 0. Because scores of hard positive pairs are clustered around 0 too, we should further suppress the score of unedited videos. Inspired by the integrated bias term of similarity normalization, we can replace it by a negative value and the similarity score with any reference videos will reduce to a negative value.

\section{Experiments}

\subsection{Implementation Details}
In video editing detection, we adopt ViT-L-16 of CLIP to extract frame features. The initial weights for RoBERTa is chinese-roberta-wwm-ext~\cite{cui-etal-2020-revisiting,chinese-bert-wwm} in huggingface.

\subsection{Results}
Our proposed method achieve outstanding performance on Meta AI Video Similarity Challenge. The results of Phase 1 and Phase 2 on Descriptor Tracks are presented in Tab.~\ref{tab:descriptor}. On Descriptor Track, our method got second place Phases 1, just 0.021 away from the first place. Although we got the first place on Phase 2, we notice that the performance drop a lot when transfer the model to Phase 2. The reason is that we only ensemble 4 models and our ensemble results are not much better than single model. Without a strong ensemble method, the transfer ability is limited. 

\begin{table}[htbp]
\centering
\scalebox{1.0}{
\begin{tabular}{l|cc}
    \toprule
    User or teams & Phase 1 $\mu$AP  & Phase 2 $\mu$AP \\
    \hline
    \cellcolor{Gray}do something(Ours) & \cellcolor{Gray}\underline{0.9176} & \cellcolor{Gray}\textbf{0.8717} \\ 
    FriendshipFirst & \textbf{0.9197} & 0.8514 \\ 
    cvl-descriptor & 0.8534 & 0.8362 \\ 
    Zihao & 0.7841& 0.7729 \\ 
    \bottomrule
\end{tabular}
}
\caption{Leaderboard results on Descriptor Track. \textbf{Bold} indicates the best result and \underline{underline} indicates the second best result.}
\label{tab:descriptor}

\end{table}

\section{Ablation Study}
To validate the effective of our proposed method, we split validation set by ourselves and analyze each module on validation set. At the beginning of the competition, we randomly divide queries in training set into 8:2 as offline training set and validation set. And the trend of performance on validation set can reflect the trend on test set. We use single basic model in ablation study because our ensemble method do not improve much. 
The results are shown in Tab.~\ref{tab:ablation}, our basic model can achieve 0.8580 on $\mu AP$, it shows that our basic model is a very strong baseline for video copy detection. Then combining frame scenes detection with basic model, the performance increased by 5\%. 
And with the video editing detection and frame scenes detection, the performance achieve 0.9492.

\begin{table}[htbp]
\centering
\scalebox{0.9}{
\begin{tabular}{l|c}
    \toprule
    Method & $\mu$AP \\
    \hline
    Basic model & 0.8580 \\
    \hline
    + FSD & 0.9075 \\
    \hline
    + VED & 0.9492 \\ 
    \bottomrule
\end{tabular}
}
\caption{Ablation study.}
\label{tab:ablation}
\end{table}

\section{Conclusion}
We introduce a dual-detection method for Video Copy Detection in this paper. The video editing detection in video-level can identify unedited videos and use random vectors with small norm and negative bias term as descriptors. The frame scenes detection in frame-level can detect scenes and split them into multiple videos, where video only have one scenes in each frame. Thought the dual-detection method, we got second place on the Descriptor Track of Meta AI Video Similarity Challenge 2022. And our descriptor give strong support to our first-place solution on Matching Track.

{\small
\bibliographystyle{ieee_fullname}
\bibliography{egbib}
}

\end{document}